\pdfoutput=1
\documentclass[11pt]{article}

\usepackage[final]{acl}

\usepackage{times}
\usepackage{latexsym}

\usepackage[T1]{fontenc}
\usepackage[utf8]{inputenc}

\usepackage{microtype}

\usepackage{inconsolata}

\usepackage{graphicx}
\usepackage{amsmath}
\usepackage{booktabs}
\usepackage{soul,sidecap}
\usepackage[nolist]{acronym}
\usepackage{amssymb}
\usepackage{pgfplots}
\pgfplotsset{compat=1.15}
\usepackage{mathrsfs}
\usetikzlibrary{arrows,decorations.pathreplacing,calligraphy}
\usepackage{tabularx,multirow,float}
\usepackage{graphicx}
\usepackage{makecell}
\usepackage{subcaption}
\usepackage[section]{placeins}

\newenvironment{myquote}{\list{}{\leftmargin=0.1in\rightmargin=0.1in}\item[]}%
  {\endlist}

\newcolumntype{C}{>{\centering\arraybackslash}p{1em}}
\newcolumntype{D}{>{\centering\arraybackslash}p{2em}}

\title{Entity or Relation Embeddings?\\ An Analysis of Encoding Strategies for Relation Extraction}

\author{Frank Mtumbuka \and Steven Schockaert \\
  Cardiff University, UK \\
  \texttt{\{MtumbukaF,SchockaertS1\}@cardiff.ac.uk}}

\begin{document}
\maketitle
\begin{abstract}
% Relation extraction is essentially a text classification problem, which can be tackled by fine-tuning a pre-trained language model (LM). However, a key challenge arises from the fact that relation extraction cannot straightforwardly be reduced to sequence or token classification. Existing approaches therefore solve the problem in an indirect way: they fine-tune an LM to learn embeddings of the head and tail entities, and then predict the relationship from these entity embeddings. Our hypothesis in this paper is that relation extraction models can be improved by capturing relationships in a more direct way. In particular, we experiment with appending a prompt with a [MASK] token, whose contextualised representation is treated as a relation embedding. While, on its own, this strategy significantly underperforms the aforementioned approach, we find that the resulting relation embeddings are highly complementary to what is captured by embeddings of the head and tail entity. By jointly considering both types of representations, we end up with a simple model that outperforms the state-of-the-art across several relation extraction benchmarks.\footnote{Our code, dataset and evaluation scripts will be made available upon acceptance.}

%Relation extraction is essentially a text classification problem. However, a key challenge arises from the fact that \steven{this task} cannot straightforwardly be reduced to sequence or token classification. 
Existing approaches to relation extraction obtain relation embeddings by concatenating embeddings of the head and tail entities. 
Despite the popularity of this approach, we find that such representations mostly capture the types of the entities involved, leading to false positives and confusion between relations that involve entities of the same type. Another possibility is to use a prompt with a [MASK] token to directly learn relation embeddings, but this approach tends to perform poorly. 
We show that this underperformance comes from the fact that information about entity types is insufficiently captured by the [MASK] embeddings. We therefore propose a simple model, which combines such [MASK] embeddings with entity embeddings. Despite its simplicity, our model consistently outperforms the state-of-the-art across several benchmarks, even when the entity embeddings are obtained from a pre-trained entity typing model. We also experiment with a self-supervised pre-training strategy which further improves the results.\footnote{Our implementation and pre-trained models are available at \url{https://github.com/fmtumbuka/RelationEmbeddings}}
\end{abstract}

\section{Introduction}
\label{sec:introduction}
Relation extraction consists in identifying the relationships between entities that are expressed in text. It is a fundamental Natural Language Processing (NLP) task, which enables the learning of symbolic representations such as knowledge graphs. While \acp{LLM} are highly effective in interpreting natural language inputs, the state-of-the-art in relation extraction is still based on fine-tuned language models of the BERT family \cite{devlin-etal-2019-bert}. Moreover, relation extraction is often applied to large document collections, such as corpora of news stories, scientific articles or social media, which means that relying on \acp{LLM} is often not feasible in practice, due to their high cost. The study of how smaller LMs can be most effectively used for this task thus remains important.

In high-level terms, the standard approach is to fine-tune a BERT-based language model  to learn a \emph{relation embedding}, i.e.\ a vector representation of the relationship that is expressed between two entities in a given sentence, and to train a classifier to predict a discrete \emph{relation label} from this embedding. To obtain these relation embeddings, we cannot simply train a sentence embedding model, since a single sentence may express several relationships. Moreover, we cannot easily identify which tokens in the sentence express the relationship, which makes learning relation embeddings fundamentally different from learning embeddings of text spans. 
In their seminal work, \citet{baldini-soares-etal-2019-matching} proposed to encapsulate the head and tail entity with special tokens. Consider the following example, where we are interested in extracting the relationship between Paris and France:
\begin{myquote}
\footnotesize
The Olympics will take place in [E1] Paris [/E1], the capital of [E2] France [/E2].
\end{myquote}
The corresponding relation embedding is obtained by concatenating the final-layer embedding of the special tokens [E1] and [E2]. This strategy was found to outperform other alternatives by \citet{baldini-soares-etal-2019-matching}, and has remained the most popular approach for learning relation embeddings %(although \citet{zhou-chen-2022-improved}, among others, have used type-specific special tokens, which encode the semantic type of the entities). 

The success of this strategy is somewhat surprising: [E1] and [E2] are intuitively designed to represent the head and tail entities, rather than their relationship even though their contextualised representations may still capture the relational context to some extent. In practice, however, knowing the semantic types of the entities often allows us to ``guess'' the relationship between them, especially if the types are fine-grained. For instance, knowing that the head entity is a capital city and the tail entity is a country, we may reasonably assume that the relationship being expressed is \emph{capital of}. 
% Nonetheless, we can expect that approaches based on entity embeddings may struggle to distinguish between closely related relations, especially if these relations involve entities of the same type. 
However, as our analysis in this paper shows, such approaches have at least two important limitations. First, they often struggle to distinguish relations between entities of the same type. Second, they sometimes lead to false positives (e.g.\ incorrectly assuming that a sentence mentioning a country and a capital city expresses the capital-of relationship).
A possible alternative is to add a prompt which includes the [MASK] token, e.g.:
\begin{myquote}
\footnotesize
The Olympics will take place in Paris, the capital of France. The relation between Paris and France is [MASK].
\end{myquote}
% We can then simply fine-tune the LM such that the contextualised embedding of the [MASK] token corresponds to a relation embedding. 
We can then fine-tune the \ac{LM} such that the contextualised embedding of [MASK] corresponds to a relation embedding.
This strategy is popular for zero-shot and few-shot relation extraction \cite{DBLP:conf/cikm/GenestPEG22}, but is not normally used for the standard supervised relation extraction setting. 
% Our experiments indeed confirm that this strategy, on its own, performs poorly. Our hypothesis is that this strategy is hampered by the fact that it cannot adequately model the semantic types of the entities, which makes relation prediction much harder. The limitations of this strategy are thus complementary to those of the more common entity embedding based strategy: in one case the model almost exclusively focuses on entity types while in the other case it largely ignores them.
% \steven{Our experiments indeed confirm that this approach performs poorly when used alone. We hypothesise that this strategy struggles because it is not able to accurately characterise the semantic types of the entities, which makes relation prediction much harder. We therefore believe that this approach is complementary to the more popular entity embedding-based strategy: in one case, the model nearly entirely concentrates on entity types, whereas in the other case, it mostly overlooks them.}
Our experiments indeed confirm that this approach performs poorly when used alone. Crucially, we find that this strategy struggles because it is not able to accurately characterise the semantic types of the entities, which makes relation prediction much harder. This approach is thus complementary to the entity embedding based strategy: in one case, the model nearly entirely concentrates on entity types, whereas in the other case, it mostly overlooks them. Exploiting this fact, we show that by combining both approaches, we arrive at a simple but highly effective strategy for relation extraction which improves the state-of-the-art in several relation extraction benchmarks. Our main contributions can be summarised as follows:
\begin{itemize}
\item We introduce a hybrid strategy which combines the entity embedding and mask based approaches, and we empirically demonstrate its surprising effectiveness.
\item We present an analysis of the entity embedding and mask based strategies, showing that the former mostly capture the entity types while the latter do not capture the entity types to a sufficient extent. Inspired by this, we experiment with a variant in which entity embeddings from a pre-trained entity typing model \cite{mtumbuka2023encore} are used instead of entity embeddings that were trained for relation extraction. Surprisingly, we find that does not deteriorate the results. 
\item Since the quality of entity and relation embeddings crucially depends on having access to sufficient training data, we also experiment with a self-supervised pre-training strategy and show that this strategy allows us to further improve the performance of all variants.
\end{itemize}

%We furthermore  show that replacing the head and tail embeddings with those from a pre-trained entity typing model \cite{mtumbuka2023encore}, we can match (or even outperform) the approach from \citet{baldini-soares-etal-2019-matching}, which clearly supports our main hypothesis that standard relation extraction methods mostly focus on semantic types.  As a final contribution, we also propose the use of a self-supervised pre-training strategy, which brings further performance gains.

% Mask not enough entity types, HT only entity types

%\frank{
%In brief, the main contributions of our work include:
%\begin{itemize}
%    \item We introduce a hybrid strategy that merges contextualized entity embeddings with relation-specific embeddings derived from a prompt-based method. We show that these embeddings are complementary.
%    \item A comprehensive comparative analysis of various encoding strategies for relation extraction. \todo{highlight EnCore results + highlight analysis that shows limitations of [HT] and Mask approaches.}
%   \item We demonstrate the effectiveness of relation embedding strategies across multiple benchmark datasets, coupled with in-depth analysis and interpretation of experimental results.
%\end{itemize}
%}

\section{Related Work}
\label{sec:related_work}

\paragraph{Learning Relation Embeddings}
The standard approach for relation extraction with LMs uses special tokens to indicate the head and tail entities (also known as the subject and object). Such approaches then predict relation labels from the contextualised representations of these two entities. For instance, the matching-the-blanks model \cite{baldini-soares-etal-2019-matching} encapsulates the head entity using the special tokens \texttt{[E1]}...\texttt{[/E1]} and the tail entity using separate special tokens \texttt{[E2]}...\texttt{[/E2]}. LUKE \cite{yamada-etal-2020-luke} simply replaces the entities by the special tokens \texttt{[HEAD]} and \texttt{[TAIL]}, omitting the actual entity spans from the input. \citet{wang-etal-2021-k} encapsulate the head and tail entity with the markers @...@ and  \#...\#, thus avoiding the introduction of new tokens. Some approaches use typed markers, which encode the semantic type of the entity, either as special tokens \citet{zhong-chen-2021-frustratingly} or by verbalising the entity type as part of the input. It should be noted, however, that entity types are typically not available in practice, which limits the applicability of such approaches.

Another possibility is to append the input with a prompt containing the [MASK] token. This strategy is popular for zero-shot and few-shot relation extraction \cite{DBLP:journals/corr/abs-2112-04539,DBLP:conf/www/ChenZXDYTHSC22,DBLP:conf/cikm/GenestPEG22}. Rather than training a classifier on top of a relation embedding, the aim is then to compare the contextualised representation of the MASK token with verbalisers, i.e.\ tokens from the LM's vocabulary that describe the relationship. 
% \steven{This strategy has rarely been considered in the fully supervised setting, with KnowPrompt \cite{DBLP:conf/www/ChenZXDYTHSC22} being a notable exception.} 
This strategy has rarely been considered in the fully supervised setting, with KnowPrompt \cite{DBLP:conf/www/ChenZXDYTHSC22} being a notable exception.

\citet{zhong-chen-2021-frustratingly} already discussed the idea that representing entity types is not sufficient for relation classification. They highlight in particular that jointly training an entity typing and relation extraction system hurts performance, suggesting that both tasks need different kinds of latent representations. We take this idea further, based on the hypothesis that representing relations by concatenating the embeddings of the head and tail entity is inherently limited, even if the entity encoders are specifically trained for relation extraction.

%************************************************
\paragraph{Pre-training Relation Encoders}
Several approaches have been proposed for pre-training or adapting language models to make them more suitable for the task of relation extraction. The matching-the-blanks model \cite{baldini-soares-etal-2019-matching} uses entity linking to find sentences that refer to the same entities, and then pre-trains a relation encoder based on the idea that sentences mentioning the same entity pairs are likely to express the same relationship. More recently, variants of this approach based on distant supervision have also been considered. Two sentences are then assumed to have the same relation if their entity pairs are asserted to have the same relation in some knowledge base, a strategy that has a long tradition in relation extraction \cite{mintz-etal-2009-distant}. For instance, \citet{peng-etal-2020-learning} implement this strategy using contrastive learning with the InfoNCE loss. While the aforementioned approaches are focused on fine-tuning a pre-trained LM, the idea of changing the LM model itself has also been explored. For instance, SpanBERT \cite{joshi-etal-2020-spanbert} changes the standard masked token prediction task with the aim of learning better span-level representations, while LUKE \cite{yamada-etal-2020-luke} uses entity linking and distinguished entity tokens to improve the representation of entities.

\paragraph{Relation Extraction with LLMs}
LLMs such as ChatGPT perform surprisingly poorly on relation extraction benchmarks, and information extraction tasks more generally \cite{DBLP:journals/corr/abs-2305-14450}. \citet{DBLP:journals/corr/abs-2305-02105} discuss some of the challenges involved in using LLMs for such tasks, which include the difficulty in selecting suitable in-context demonstrations. 
\citet{DBLP:journals/corr/abs-2311-08993} make the observation that LLMs with in-context learning (ICL) struggle in particular on specification-heavy tasks, i.e.\ tasks where even human annotators need to carefully study a non-trivial set of annotation guidelines to correctly solve the task, as is often the case in information extraction. Due to the challenges of using ICL for relation extraction, most approaches involving LLMs use models that can be fine-tuned. For instance, \citet{sainz-etal-2021-label} proposed a reformulation of relation extraction as a Natural Language Inference (NLI) problem and included experiments with the 1.5B parameter DeBERTa\textsubscript{XXL} model \cite{DBLP:conf/iclr/HeLGC21}. While this allowed them to improve the state-of-the-art at the time, it should be noted that the NLI based formulation is highly inefficient when a large number of relation labels need to be considered. It also relies on manually defined verbalisations of the relation labels. \citet{DBLP:journals/corr/abs-2205-10475} fine-tuned a 10B parameter model on a range of tasks that can be formulated as triple prediction, including relation extraction. \citet{wadhwa-etal-2023-revisiting} found that while GPT-3 \cite{DBLP:conf/nips/BrownMRSKDNSSAA20} performed poorly when used directly, it was useful for generating chain-of-thought \cite{DBLP:conf/nips/Wei0SBIXCLZ22} explanations. In particular, they showed that by fine-tuning a Flan-T5 \cite{DBLP:journals/corr/abs-2210-11416} model on these explanations, the resulting model performed substantially better than when fine-tuning Flan-T5 on the relation extraction task directly. As another strategy for leveraging LLMs indirectly, \citet{DBLP:conf/sustainlp/XuZWZ23} use ChatGPT for data generation in few-shot settings. Overall, however, the state-of-the-art in relation extraction, for the fully supervised setting, is still based on fine-tuned models of the BERT family \cite{devlin-etal-2019-bert}. While this might change in future, e.g.\ with novel prompting techniques or better models, the need for efficient information extraction models means that such smaller models are likely to remain important.

\section{Relation Extraction}
\label{sec:our_approach}
We consider the standard sentence-level relation extraction setting, where we are given a sentence in which two entities are highlighted, which we refer to as the \emph{head entity} and \emph{tail entity}. The goal is to predict which relationship holds between these two entities, given a pre-defined set of candidate relation labels. We focus on strategies that first learn a relation embedding, which describes the relationship between the two entities in a continuous space, and then use a classifier to predict the actual label based on that relation embedding. In Section \ref{subsec:pretraining_encoder}, we first explain the pre-training strategies that we use for learning high-quality relation embeddings. Section~\ref{subsec:relation_classification} then describes how the pre-trained relation encoder is fine-tuned for the relation classification task. Finally, in Section \ref{secVariants} we explain how relation embeddings can be obtained by concatenating the contextualised embeddings of the head and tail entities or by using a prompt-based strategy with a [MASK] token, among others.

%---------------------------------------------
\subsection{Pre-training the Relation Encoder}
\label{subsec:pretraining_encoder}

\paragraph{Pre-Training Objective}
To pre-train relation encoders, we rely on the InfoNCE contrastive loss \cite{DBLP:journals/corr/abs-1807-03748}, which has been found effective for learning relation embeddings \cite{peng-etal-2020-learning}, and for representation learning in NLP more generally \cite{gao-etal-2021-simcse,liu-etal-2021-mirrorwic,DBLP:conf/sigir/LiKBS23,mtumbuka2023encore}. Specifically, let us assume that we have a set $S$ of sentences with designated head and tail entities. For each $s\in S$, we assume that we have access to a set of positive examples $P_s$, i.e.\ sentences which express the same relationship as the one expressed in $s$, and a set of negative examples $N_s$. Let us write $\phi(s)$ for the relation embedding obtained from sentence $s$ using some encoding strategy. For instance, $\phi(s)$ may be the concatenation of the contextualised representations of the head and tail entity, or it may be the representation of the [MASK] token when a relation prompt is used. Section \ref{secVariants} will describe the specific encoding strategies that we consider in our analysis. We train the encoder $\phi$ using the InfoNCE loss:
\begin{align}\label{eqInfoNCE}
{-}\sum_{s\in S}\sum_{p\in P_s}\log\frac{\exp\big({\cos}(\phi(s),\phi(p))/\tau\big)}{\sum_{x} \exp\big({\cos}(\phi(s),\phi(x))/\tau\big)}
\end{align}
where the temperature $\tau>0$ is a hyperparameter, and the summation in the denominator ranges over $x\in N_s\cup \{p\}$. The loss captures the intuition that two sentences expressing the same relationship should have similar relation embeddings. As suggested by previous work \cite{baldini-soares-etal-2019-matching,peng-etal-2020-learning} we also include the masked language modelling (MLM) objective during pre-training to prevent catastrophic forgetting. The overall loss is thus given by $\mathcal{L}_{\text{info}} + \mathcal{L}_{\text{MLM}}$, with $\mathcal{L}_{\text{info}}$ the loss in \eqref{eqInfoNCE} and $\mathcal{L}_{\text{MLM}}$ the MLM objective.

\paragraph{Self-Supervised Pre-Training}
The effectiveness of the pre-training objective crucially depends on the quality and quantity of the available examples. In most cases, we have access to a set of labelled examples, obtained through manual annotation or distant supervision. The positive examples $P_s$ are then simply those examples that have the same label as $s$. 
% As an alternative, we will also experiment with a form of self-supervised pre-training, using coreference chains as a supervision signal. 
As an alternative, we also experiment with a form of self-supervised pre-training, using coreference chains as a supervision signal.
Specifically, we adapt the EnCore strategy from \citet{mtumbuka2023encore} for pre-training entity encoders, by proposing a similar strategy for learning relation embeddings. 
The central idea is that two sentences are likely to express the same relationship if they refer to the same two entities. The matching-the-blank model also relies on this idea, but uses entity linking to identify such sentence pairs. Following \citet{mtumbuka2023encore} we select positive examples from the Gigaword corpus\footnote{\url{https://catalog.ldc.upenn.edu/LDC2003T05}} and we only consider two entities to be co-referring if they are identified as such by two separate off-the-shelf coreference systems: the \textit{Explosion AI} system Coreferee v1.3.1\footnote{\url{https://github.com/explosion/coreferee}} and the \textit{AllenNLP} coreference model\footnote{\url{https://demo.allennlp.org/coreference-resolution}}. This use of two coreference systems was found to reduce the number of false positives because of spurious coreference links, given that state-of-the-art coreference resolution systems are still far from perfect. 
% \steven{Clearly, the fact that two sentences mention the same entities does not guarantee that the sentences actually express the same relationship, which is a common limitation of self-supervised strategies. Note, however, that this issue is somewhat mitigated because we only consider sentences from the same news story.}
Clearly, the fact that two sentences mention the same entities does not guarantee that the sentences actually express the same relationship, which is a common limitation of self-supervised strategies. Note, however, that this issue is somewhat mitigated because we only consider sentences from the same news story.

\subsection{Relation Classification}
\label{subsec:relation_classification}
Given a sentence $s$ with designated head and tail entities, we use the pre-trained relation encoder to obtain an embedding $\phi(s)$ that captures the relationship between these entities. Now consider the problem of classifying this relationship, using the labels from some set $\{l_1,...,l_m\}$. Following standard practice \cite{zhong-chen-2021-frustratingly,zhou-chen-2022-improved}, we use a feedforward network with one 
hidden layer and ReLU activation, i.e.\ predictions are made as follows:
\begin{align*}
\mathbf{h} &= \mathsf{ReLU}(\mathbf{A_1} \phi(s) + \mathbf{b_1})\\
(p_1,...,p_m) &= \mathsf{softmax}(\mathbf{A_2} \mathbf{h} + \mathbf{b_2})
\end{align*}
where $p_i$ is interpreted as the probability that that $l_i$ is the correct label, $\mathbf{A_1}$ and $\mathbf{A_2}$ are matrices, and $\mathbf{b_1}$ and $\mathbf{b_2}$ are bias terms. The label classifier is trained using cross-entropy. We also fine-tune the pre-trained relation encoder during this step. 
% The dimension of the hidden representations $\mathbf{h}$ is chosen to be the same as the dimension of input $\phi(s)$.
The dimension of the hidden representations $\mathbf{h}$ is set to be the same as that of the corresponding encoder. For instance, for all experiments with BERT-base, we set the hidden layer to 768 dimensions.

%*****************************************************************
\subsection{Encoding Strategies}\label{secVariants}
We now discuss the considered strategies for obtaining relation embeddings. Suppose we are interested in the relationship between the entities \textit{<h>} and \textit{<t>} in sentence $s$. We first create an annotated version of sentence $s$, where (i) \textit{<h>} is encapsulated with the special tokens \texttt{[E1]}...\texttt{[/E1]} and \textit{<t>} is encapsulated with the special tokens \texttt{[E2]}...\texttt{[/E2]}, and we append the phrase ``The relation between \textit{<h>} and \textit{<t>} is [MASK]''. For instance:
%, for the example from the introduction we obtain the following annotated sentence:
\begin{myquote}
\footnotesize
The Olympics will take place in [E1] Paris [/E1], the capital of [E2] France [/E2]. The relation between Paris and France is [MASK].
\end{myquote}
For some of our strategies the [MASK] token will not be used, while another strategy only uses the [MASK] token. However, we use the same annotated sentence in all cases, as this allows for the most direct comparison. First, we consider the following basic approaches:
\begin{description}
\item{\textbf{[H,T]}} We define $\phi(s)$ as the concatenation of the final-layer embeddings of the tokens [E1] and [E2], following \citet{baldini-soares-etal-2019-matching}.
\item{\textbf{Mask}} We define $\phi(s)$ as the final-layer representation of the [MASK] token.
\item{\textbf{[H,T,Mask]}} We define $\phi(s)$ as the concatenation of the final-layer embeddings of the tokens [E1], [E2] and [MASK].
\end{description}
In each case, we pre-train the relation encoder as explained in Section \ref{subsec:pretraining_encoder} and then train a classifier as explained in Section \ref{subsec:relation_classification}. 

\begin{figure}
\centering
\includegraphics[width=140pt]{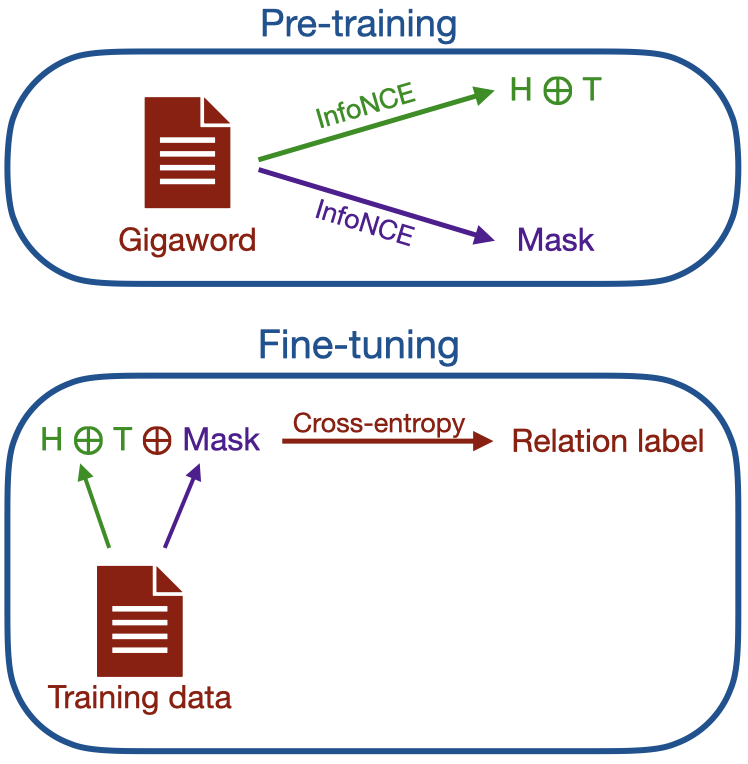}
\caption{Illustration of the [H,T]+Mask strategy.\label{figHTpM}}
\end{figure}

\begin{figure}
\centering
\includegraphics[width=140pt]{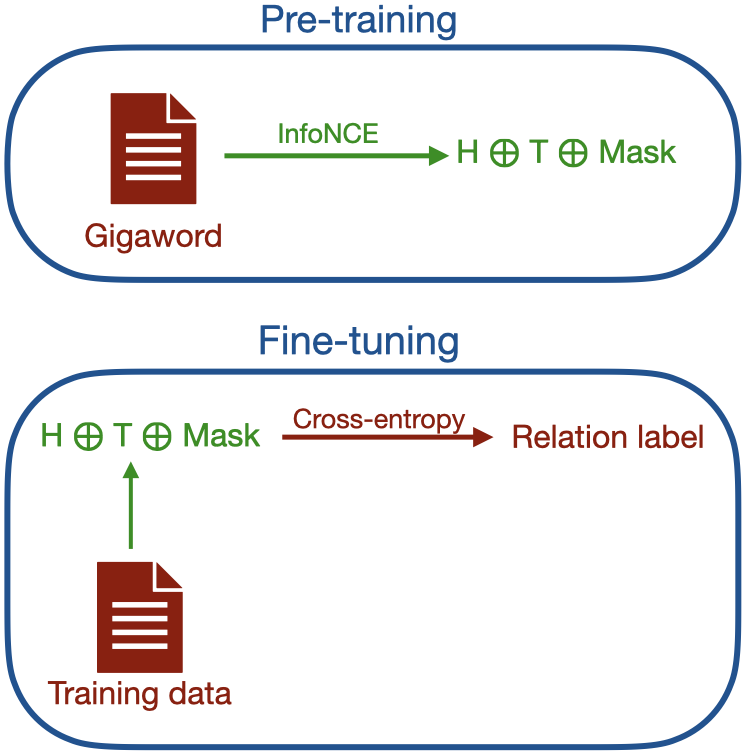}
\caption{Illustration of the [H,T,Mask] strategy.\label{figHTM}}
\end{figure}

\paragraph{Hybrid Strategy} 
% We also consider a hybrid strategy, in which two different encoders are used during pre-training: $\phi_1$, which uses the [H,T] strategy, and $\phi_2$, which uses the Mask strategy. The pre-training loss is then of the form $\mathcal{L}_{\text{info}}^1 + \mathcal{L}_{\text{info}}^2 + \mathcal{L}_{\text{MLM}}$, where $\mathcal{L}_{\text{info}}^1$ and $\mathcal{L}_{\text{info}}^2$ both correspond to the InfoNCE loss \eqref{eqInfoNCE}, but where the former refers to the encoder $\phi_1$ and the latter refers to the encoder $\phi_2$. The encoders $\phi_1$ and $\phi_2$ share the same parameters, i.e.\ we fine-tune a single language model, but either use the concatenation of the embeddings of [E1] and [E2] as the relation embedding, or the embedding of the [MASK] token. After the pre-training step, when training the classifier, we simply concatenate the embeddings of the [E1], [E2] and [MASK] tokens. We will refer to this strategy as \textbf{[H,T]+Mask}. Note that the only difference with [H,T,Mask] lies in how the encoder is pre-trained. The motivation for the [H,T]+Mask strategy comes from the idea that the [MASK] token may be largely ignored when using the [H,T,Mask] strategy, since it may be easier to learn meaningful representations of the entities than to learn meaningful relation embeddings. The hybrid training strategy [H,T]+Mask avoids this issue, by forcing the [MASK] token to capture a meaningful relation embedding, before combining this representation with the contextualised entity embeddings.
We also consider a hybrid approach, where we pre-train the relation encoder using a loss of the form $\mathcal{L}_{\text{info}}^1 + \mathcal{L}_{\text{info}}^2 + \mathcal{L}_{\text{MLM}}$. $\mathcal{L}_{\text{info}}^1$ and $\mathcal{L}_{\text{info}}^2$ both relate to the InfoNCE loss \eqref{eqInfoNCE} but refer to different representations: $\mathcal{L}_{\text{info}}^1$ uses the [H,T] representation while $\mathcal{L}_{\text{info}}^2$ uses the Mask representation. Note that we fine-tune a single language model, i.e.\ the [H,T] and Mask representations are obtained with the same encoder. After the pre-training step, when training the classifier, we concatenate the two representations (i.e.\ the embeddings of the [E1], [E2] and [MASK] tokens). We will refer to this strategy as \textbf{[H,T]+Mask}. Note that the only difference with [H,T,Mask] lies in how the encoder is pre-trained. The underlying motivation comes from the idea that the [MASK] token may be largely ignored when using the [H,T,Mask] strategy, since it may be easier to learn meaningful representations of the entities than to learn meaningful relation embeddings. The hybrid training strategy [H,T]+Mask avoids this issue, by forcing the [MASK] token to capture a meaningful relation embedding, before combining this representation with the contextualised entity embeddings.  Figures \ref{figHTpM} and \ref{figHTM} illustrate the difference between the [H,T,Mask] and [H,T]+Mask strategies.

\paragraph{Pre-trained Entity Embeddings} Our main hypothesis is that the [H,T] strategy focuses on learning the semantic types of the head and tail entities. To test this hypothesis, we consider a variant in which we use an entity embedding model instead of pre-training a relation encoder using the [H,T] strategy. In particular, we rely on EnCore \cite{mtumbuka2023encore} as the pre-trained entity embedding model. The EnCore embeddings essentially capture the semantic types of the entities (but without reference to any particular set of type labels). Crucially, no relational knowledge is used during the EnCore training process. We consider the following variants:
\begin{description}
\item[\textbf{EnCore}] No pre-training is used. We directly train a classifier on the concatenation of the EnCore embeddings of the head and tail entities.
\item[\textbf{EnCore+Mask}] We pre-train an encoder using the Mask strategy. Then we train the classifier on the concatenation of the [MASK] token and the EnCore entity embeddings.
\item[\textbf{EnCore+[H,T]+Mask}] We use the same hybrid pre-training as the [H,T]+Mask strategy. The classifier is then trained on the concatenation of the [H,T]+Mask representation and the EnCore entity embeddings.
\end{description}
Note that the EnCore model itself is not fine-tuned. This ensures that the embeddings provided by this model remain focused on entity types. 

\section{Experiments}
\label{sec:experimental_analysis}
We compare the effectiveness of the considered relation embedding strategies on a number of standard relation extraction benchmarks. Our main hypothesis is that the common [H,T] strategy essentially leads models to focus on the semantic types of the head and tail entity. We are thus interested in comparing the [H,T] and EnCore strategies. Furthermore, we hypothesise that the information captured by the Mask embeddings is complementary to that captured by the [H,T] embeddings. Accordingly, we are interested in comparing [H,T] with [H,T,Mask] and [H,T]+Mask.
%We first present the details of our experimental setup in Section \ref{subsec:experimental_setup}. Section \ref{subsec:results} subsequently summarises our main results. Finally, we present some additional analysis in Section \ref{sec:analysis}.

%************************************************************
\subsection{Experimental Setup}
\label{subsec:experimental_setup}

\paragraph{Benchmarks}
We evaluate on five standard benchmarks. First, we use TACRED~\cite{zhang-etal-2017-position}, a popular relation extraction benchmark. Two revisions of this dataset have been proposed, both of which are aimed at addressing problems with noisy annotations. In particular, TACREV~\cite{alt-etal-2020-tacred} was obtained by re-annotating the 5000 most challenging instances from the development and test sets. Specifically, to select these instances, the authors looked at how often models disagreed with each other and with the ground truth. Re-TACRED~\cite{Stoica_Platanios_Poczos_2021} was obtained by re-annotating the entire dataset. Following tradition, we report results on all three variants in terms of F1 score. Next, we evaluate on a distantly supervised dataset that was introduced by \citet{sorokin-gurevych-2017-context} by aligning Wikipedia and Wikidata (Wiki-WD). 
%This is a large dataset, involving 353 relations classes, 372,059 training sentences in training and 360,334 for test sentences. 
Finally, we also consider the distantly supervised dataset that was introduced by \citet{nyt_freebase} by aligning articles from the New York Times with Freebase (NYT-FB).
The NYT-FB dataset does not have an explicit validation set. As a result, we keep 10\% of the training set as a validation set and train on the remaining 90\%.
Following the tradition from previous work, we report the results on Wiki-WD in terms of F1 score and the results on NYT-FB in terms of precision at 10 (P@10) and 30 (P@30), averaged across all relation labels.
%Note that while the Wiki-WD benchmark contains sentences that express multiple relationships, the NYT-FB dataset only contains sentences that express a single relationship.
%This dataset covers 53 relation classes,  455,771 sentences for training and 172,448 sentences for testing. 
Table \ref{tabOverviewDatasets} summarises the main characteristics of the considered datasets.

\label{subsec:benchmarks}
\begin{table}
    \footnotesize
    \setlength\tabcolsep{4pt}
    \centering
    \begin{tabular}{lcccc}
        \toprule
        \textbf{Dataset} & \textbf{\# Class} & \textbf{Train} & \textbf{Dev.} & \textbf{Test}\\
        \midrule
        TACRED & 42 & 68.1K & 22.6K & 15.5K \\
        TACREV & 42 & 68.1K & 22.6K & 15.5K\\
        ReTACRED & 40 & 58.4K & 19.5K & 13.4K\\
        Wiki-WD & 353 & 372.1K & 123.8K & 360.3K\\
        NYT-FB & 53 & 455.8K & - & 172.4K\\
        \bottomrule
    \end{tabular}
    \caption{Overview of the considered benchmarks, showing the number of distinct relation classes, and the number of annotated mentions in the training, development and test sets. \label{tabOverviewDatasets}}
\end{table}

%*****************************
\paragraph{Baselines}
FOR TACRED and its variants, we consider a number of recent baselines. First, we include a comparison with SpanBERT \cite{joshi-etal-2020-spanbert} and LUKE \cite{yamada-etal-2020-luke}. We furthermore compare with KnowBERT \cite{DBLP:conf/www/ChenZXDYTHSC22}, which also uses a prompt with the [MASK] token. They improve on the standard Mask strategy, among others, by incorporating predicted entity types, and are thus a natural baseline for our methods.
Finally, we consider the Typed Marker strategy from   \citet{zhou-chen-2022-improved}, and the Curriculum Learning variant from \citet{curriculum_learning}. Note, however, that these last two methods are not directly comparable with our methods, as they rely on the gold entity type labels that are provided as part of the TACRED dataset. We do not use these labels for our models since such information is typically not available in practice. 
For Wiki-WD and NYT-FB, we compare with RECON~\cite{recon} and KGPool~\cite{nadgeri-etal-2021-kgpool}. Note, however, that these state-of-the-art methods are again not directly comparable with our methods. They are focused on modelling and exploiting knowledge from the Wikidata and Freebase knowledge graphs. We do not take such information into account, as our focus is on comparing different encoding strategies for learning relation embeddings. 

%*****************************
\paragraph{Training Strategies}
Unless stated otherwise, the relation encoder is first pre-trained using the strategy from Section \ref{subsec:pretraining_encoder}, before being fine-tuned, as explained in Section \ref{subsec:relation_classification}. In our default setting, we use the same training set for pre-training and fine-tuning the relation encoder. For TACRED, TACREV and ReTACRED, where the training set is comparatively smaller, we also experiment with a variant where we instead use the Gigaword corpus for pre-training the relation encoder, following the self-supervision strategy from Section \ref{subsec:pretraining_encoder}. Finally, we also report results where the pre-training  step is omitted and we directly train the relation encoder using the fine-tuning strategy from Section \ref{subsec:relation_classification}.
For our main experiments, we use {\texttt roberta-large}\footnote{\url{https://huggingface.co/docs/transformers/model_doc/roberta}} to initialise the relation encoder.

%***********************************
\subsection{Results}
\label{subsec:results}

  \begin{table}[t]
 	\begin{center}
 		\footnotesize{
 			\centering{
 	    \begin{tabular}{llll}
 		\toprule
        &  \multicolumn{1}{c}{\rotatebox{90}{\textbf{TACRED}}}
        &  \multicolumn{1}{c}{\rotatebox{90}{\textbf{TACREV}}}
        &  \multicolumn{1}{c}{\rotatebox{90}{\textbf{Re-TACRED}}}\\ 	    
   \midrule
      \multicolumn{4}{c}{\textit{Baselines}}\\
    \midrule
    SpanBERT    & 70.8 & 78.0$^*$ & 85.3$^{\dagger}$  \\
    LUKE & 72.7  & 80.6$^{\circ}$ & 90.3$^{\circ}$\\
    KnowPrompt   & 72.4 & 81.4 & 90.9\\
    Typed Marker    & 74.6 & 83.2 & 91.1  \\
    Curriculum Learning    & 75.0 & -  & 91.4  \\
    \midrule
    \multicolumn{4}{c}{\textit{Standard pre-training}}\\
    \midrule
    \text{Mask} & 23.3 & 22.9 & 23.2 \\
    \text{[H,T]} & 61.4 & 63.6 & 72.3 \\
    \text{[H,T,Mask]} & 73.0 & 73.8 & 81.9\\
    \text{[H,T]+Mask} &  78.5 & 84.6 &  91.9  \\
    EnCore+Mask  & 78.9 & 83.9 &   91.8  \\
    \text{EnCore+[H,T]+Mask}  & 78.1  & 84.1 & 91.8  \\
    \midrule
    \multicolumn{4}{c}{\textit{Gigaword pre-training}}\\
    \midrule
    \text{Mask} & 24.6 & 23.5 & 38.2 \\
    \text{[H,T]} & 63.2 & 66.3 & 79.7 \\
    \text{[H,T,Mask]} & 74.4 & 75.1 & 82.9\\
    \text{[H,T]+Mask} & 78.5 & 83.6 & 92.7 \\
    EnCore+Mask  & \textbf{79.1} & \textbf{84.9} &   \textbf{93.5}  \\
    \text{EnCore+[H,T]+Mask} & 79.0  & 84.4 & 93.2  \\
    \midrule
    \multicolumn{4}{c}{\textit{No pre-training}}\\
    \midrule
    \text{[H,T,Mask]}  & 55.2 & 54.3 & 60.4 \\  
    EnCore & 69.8 & 76.1 & 80.4 \\
    \bottomrule                                
     		\end{tabular}}}
 	\caption{Comparison of different relation embedding strategies, in terms of F1 ($\%$). Results marked with $*$ were taken from \citet{alt-etal-2020-tacred}, those marked with $\dagger$ were taken from \citet{Stoica_Platanios_Poczos_2021}, and those marked with $\circ$ were taken from \citet{zhou-chen-2022-improved}. All other baseline results were taken from the original papers. Our models are initialised from RoBERTa-large, which is the same for all baselines except for SpanBERT, which uses BERT-large.}
 	 \label{tab:comparisons_to_baselines_tacred}
 	\end{center}
 \end{table}

\begin{table}[t]
 	\begin{center}
 		\footnotesize
 			\centering 
            \setlength\tabcolsep{3.5pt}
     	    \begin{tabular}{lccccc}
     		\toprule
            & \multicolumn{3}{c}{\textbf{Wiki-WD}} 
            & \multicolumn{2}{c}{\textbf{NYT-FB}}\\
     	    \cmidrule(lr){2-4}
            \cmidrule(lr){5-6}
            & \textbf{P} & \textbf{R} & \textbf{F1} & \textbf{P@10} & \textbf{P@30}\\
       \midrule
       \multicolumn{6}{c}{\textit{Baselines}}\\
       \midrule
        RECON&  87.2 & 87.2 & 87.2 & 87.5 & 74.1\\
        KGPool & 88.6 & 88.6 & 88.6 & 92.3 & 86.7\\
        \midrule
       \multicolumn{6}{c}{\textit{Standard pre-training}}\\
       \midrule
        Mask & 50.3 & 48.7 & 48.9 & 52.3 & 49.7\\
        \text{[H,T]}  & 75.9 & 74.7 & 75.2 & 79.6 & 78.5\\
        \text{[H,T,Mask]}  & 82.1 & 80.8 & 81.7 & 88.4 & 86.1\\
        \text{[H,T]+Mask} &  \textbf{89.8} & \textbf{89.3} & \textbf{89.6} & 94.7 & \textbf{93.1}\\
        EnCore+Mask  & 89.4 & 88.8 & 89.1 & \textbf{94.9} & 93.0\\
        \text{EnCore+[H,T]+Mask} & \textbf{89.8} & 88.6 & 88.9 & 94.7 & 92.9\\
        \midrule
       \multicolumn{6}{c}{\textit{No pre-training}}\\
       \midrule
        \text{[H,T,Mask]} &  56.1 & 54.7 & 55.1 & 58.1 & 56.9\\
        EnCore  & 80.8 & 79.2 & 79.8& 86.1 & 84.7\\
        \bottomrule                                
         		\end{tabular}
     	\caption{Comparison of different relation embedding strategies. Baseline results were obtained from the original papers. Our models are initialised from RoBERTa-large.}\label{tab:comparisons_to_baselines_wikidata}%
     	\end{center}
     \end{table}

%\begin{table}[t]
%\centering
%\footnotesize
%\begin{tabular}{lccccc}
%\toprule
%& & \rotatebox{90}{\textbf{BERT-base}} & \rotatebox{90}{\textbf{BERT-large}} & \rotatebox{90}{\textbf{RoBERTa-large}} & \rotatebox{90}{\textbf{ALBERT-xxl}} \\
%\midrule
%TACRED & F1 & 74.8 & 75.1 & 78.5 & \textbf{78.9} \\
%TACREV & F1 & 75.6 & 77.3 & 84.6 & \textbf{83.9} \\
%Re-TACRED & F1 & 86.4 & 87.9 & 91.9 & \textbf{92.3} \\
%Wiki-WD & F1 & 79.7 & 84.9 & 89.6 & \textbf{89.8} \\
%NYT-FB & P@10 & 93.9 & 94.1 & 94.7 & \textbf{94.9} \\
%NYT-FB & P@30 & 93.4 & 92.5 & 93.1 & \textbf{93.8} \\
%\bottomrule
%\end{tabular}
%\caption{Comparison of different language models, using the [H,T]+Mask relation embedding strategy under standard pre-training.}
%\label{tabComparisonLMs}
%\end{table}

 \begin{table*}[t]
    \centering
    \footnotesize
    \setlength\tabcolsep{4pt}
    \begin{tabular}{p{285pt}p{150pt}}
        \toprule
         \textbf{Sentence} 
         & \textbf{Label}\\ 
         \midrule
         \textbf{Asia Bibi} was sentenced to hang in Pakistan's central province of \textbf{Punjab} earlier this month after being accused of insulting the Prophet Mohammed in 2009.
         & Gold: no relation \newline Mask: per:origin \newline [H,T]: per:state or provinces of residence\newline [H,T]+Mask: no relation\\
         \midrule          
         %\textbf{John Graham}, a 55-year-old man from \textbf{Canada}, is accused of shooting Aquash in the head and leaving her to die on the Pine Ridge reservation in South Dakota.
         %& Gold: per:origin\newline [H,T]: per:countries of residence\newline [H,T]+Mask: per:origin\\
         %\midrule
       %  ``To get rich is glorious" has been the mantra in booming \textbf{communist China} for 30 years, but few have embraced the slogan more vigorously than \textbf{Wen Qiang,} a leading law official in the country's southwest.
        % & Gold: no relation \newline Mask: per:origin \newline [H,T]: per:countries of residence\newline [H,T]+Mask: no relation\\
         In October, \textbf{she} filed a complaint with the police in \textbf{Rio} saying he had kidnapped her and tried to threaten her into having an abortion.
         & Gold: no relation \newline Mask: per:origin \newline [H,T]: per:cities of residence \newline [H,T]+Mask: no relation \\
         \midrule
         \textbf{Benjamin Chertoff} is the Editor in Chief of Popular Mechanics magazine as well as the cousin of the Director of \textbf{Homeland Security}, Michael Chertoff.
         & Gold: no relation \newline Mask: per:employee of \newline [H,T]: per:employee of\newline [H,T]+Mask: no relation\\
         \midrule
         WASHINGTON -- \textbf{The National Restaurant Association} gave \$35,000 -- a year's salary -- in severance pay to a female staff member in the late 1990s after an encounter with \textbf{Herman Cain}, its chief executive at the time, made her uncomfortable working there, three people with direct knowledge of the payment said on Tuesday.
         & Gold: org:top members/employees \newline Mask: per:employee of \newline [H,T]: no relation\newline [H,T]+Mask: org:top members/employees\\
         % \midrule
         % This news comes from Karr Ingham, an economist who created the \textbf{Texas Petro Index -LRB- TPI -RRB-}, which is a service of the \textbf{Texas Alliance of Energy Producers}.
         % & Gold: org:parents \newline Mask: no relation \newline [H,T]: no relation \newline [H,T]+Mask: org:parents \\
         % \midrule
         % ``Just as the Japanese learned to make cars in America without Japanese workers, Indian vendors are learning to outsource without Indians," said Dennis McGuire, chairman of \textbf{TPI}, a consultancy based in \textbf{Texas}, that focuses on outsourcing.
         % & Gold: org:state or province of headquarters \newline Mask: per:origin \newline [H,T]: no relation  \newline [H,T]+Mask: org:state or province of headquarters \\
     \midrule
         ``From January 1, I,\textbf{ Charles Ble Goude} and the youth of Ivory Coast are going to liberate the Golf Hotel with our bare hands," the political showman turned \textbf{minister} declared Wednesday, to a cheering crowd of hardline supporters.
         & Gold: per:title \newline Mask: per:employee of \newline [H,T]: no relation \newline [H,T]+Mask: per:title \\
   % \midrule
   %      He figured that he would sell his home before the interest rate on the loan, taken out from \textbf{Countrywide Financial}, now owned by \textbf{Bank  of America}, reset at a higher level.
   %      & Gold: org:parents \newline Mask: org:alternate names  \newline [H,T]: no relation \newline [H,T]+Mask: no relation \\
    \bottomrule
    \end{tabular}
    % \vspace*{-2ex}
    \caption{Comparison of the TACREV test set predictions from the Mask, [H, T] and [H, T] + Mask models that were initialised using RoBERTa-large.}
    \label{tabComparisonModelOutputs}
\end{table*}

Table \ref{tab:comparisons_to_baselines_tacred} summarises the results for TACRED, TACREV and Re-TACRED. Let us first consider the standard pre-training results (i.e.\ where the models are pre-trained on the standard training set). A number of observations stand out. 
First, the Mask strategy on its own performs poorly.
%\frank{First, the Mask strategy is not very effective by itself. This is also true when compared to the baseline, which just takes into account the CLS token, particularly on the Tacred-family datasets.}
Second, there is clear evidence that the [H,T] and Mask strategies are complementary: both [H,T,Mask] and [H,T]+Mask substantially outperform [H,T] and Mask on their own. We can furthermore see that [H,T]+Mask outperforms [H,T,Mask]. The combined pre-training strategy which is used by [H,T,Mask] thus indeed seems to largely fail to learn meaningful [MASK] embeddings. Finally, the EnCore+Mask strategy matches or even outperforms [H,T]+Mask. This is surprising, given that the EnCore embeddings are specifically trained to capture semantic types and are not fine-tuned on the relation extraction datasets. The strong performance of EnCore+Mask thus clearly supports the idea that the [H,T] representations mostly capture the entity types, rather than the relationship itself. We can similarly see that EnCore+[H,T]+Mask does not generally improve on Encore+Mask, which further supports this idea.

Pre-training on Gigaword leads to clear and consistent improvements, compared to standard pre-training.
%the effectiveness of this pre-training strategy, as the results of all models are better than with standard pre-training. 
While the fact that pre-training on external datasets can bring benefits is in itself unsurprising, this self-supervision strategy based on coreference chains was not previously tested for relation extraction. It offers a convenient way to improve relation extraction systems, since it does not rely on an entity linked corpus. Comparing the performance of the different configurations, after pre-training on Gigaword, we see the same patterns as with standard pre-training. The Encore+Mask strategy emerges as the best model overall, which in particular confirms the usefulness of combining entity embeddings information with relation embeddings.

We can also see that forgoing pre-training altogether has a detrimental effect. A direct comparison with the baselines is difficult, as the two strongest baselines use additional information (i.e.\ the gold entity type labels). Nonetheless, our best configurations consistently outperform the state-of-the-art methods, with the improvements being most pronounced for TACRED. %The improvements are relatively small for TACREV and Re-TACRED, especially with standard pre-training, while being somewhat larger for TACRED. 

Table \ref{tab:comparisons_to_baselines_wikidata} summarises the results we obtained for Wiki-WD and NYT-FB. The main patterns are consistent with the results from Table \ref{tab:comparisons_to_baselines_tacred}. For instance, we can again see that Mask on its own performs poorly and that [H,T]+Mask outperforms both Mask and [H,T] by a considerable margin. We furthermore again see that the pre-trained entity embeddings from EnCore can serve the same purpose as the fine-tuned [H,T] embeddings. The best configurations outperform the baselines, with the improvements on NYT-FB being particularly clear. However, these methods are not directly comparable as they focus on different information.

Further analysis of our results can be found in the appendix. Among others, we show that the conclusions from this section remain valid when other language models than RoBERTa-large are used as the encoder. We also present a detailed error analysis to support our claims about the limitations of the [H,T] and Mask strategies, an analysis of the models in a setting with limited training data, and an analysis of the impact of the dimensionality of the entity and relation embeddings.

%********************************************************
\subsection{Qualitative Analysis}\label{sec:analysis}
%\paragraph{Qualitative Analysis of Model Outputs}
In Table~\ref{tabComparisonModelOutputs}, we compare predictions of the Mask, [H,T] and [H,T]+Mask models for the TACREV test set. 
The first three cases illustrate how the [H,T] model often overly relies on the semantic types of the entities, without fully taking into account the actual sentence context, a problem known as entity bias \cite{wang-etal-2022-rely}. In particular, as the first two examples illustrate, given a person and a place, the [H,T] model frequently predicts that the individual is a resident of that place, even if there is no relationship expressed in the given context. In contrast, [H,T]+Mask correctly predicts \textit{no relation} in these cases. The third example shows a similar issue, which arises when the sentence refers to a person and an organisation. In this case, the [H,T] model incorrectly predicts that the person is employed by that organisation. As the fourth example illustrates, however, the opposite situation also arises, where the [H,T] model incorrectly predicts \textit{no relation}. Furthermore, several examples illustrate how the Mask model struggles because it does not adequately capture the semantic types of the entities. For instance, in the first two examples, the model predicts \textit{per:origin} despite the fact that the tail entity is not a country. The issue is most clearly illustrated by the fifth example, where the Mask model predicts \textit{employee of}, despite the tail entity not being an organisation.
%The last example illustrates a case where all models make an incorrect prediction. 
Overall, these examples support the view that the [H,T] model focuses too much on modelling the entity types, which is not always sufficient, while conversely, Mask struggles because it does not sufficiently take entity types into account. We include further examples in the appendix, which further support our findings.

%where we also present an analysis of the confusion matrices to provide more insights.

%These cases illustrate how the Mask and [H,T] models often overly rely on the semantic types of the entities, without fully taking into account the actual sentence context, a problem known as entity bias \cite{wang-etal-2022-rely}.As the first two examples illustrate, given a person and a country, the Mask and [H,T] models frequently predict that the individual is either a native of that country or currently resides there, even if there is no relationship expressed in the given context. \steven{In contrast, [H,T]+Mask correctly predict \textit{no relation} in these cases.} The third example shows a similar issue, which arises when the sentence refers to a person and an organisation. \steven{In this case, both the} Mask and [H,T] models incorrectly predict that the person is employed by that organisation. As the fourth example illustrates, however, the opposite situation also arises, where \steven{the [H,T] model incorrectly predicts \textit{no relation}. The last example illustrates a case where all models make an incorrect prediction.} These examples support the view that the [H,T] model focuses almost exclusively on modelling the entity types, which is not always sufficient, and Mask alone does not learn meaningful relation embeddings. \steven{We include further examples in the appendix, where we also present an analysis of the confusion matrices.}

\section{Conclusions}
\label{sec:conclusion}
The primary aim of this paper was to analyse two different strategies for training relation encoders. On the one hand, most work in supervised relation extraction relies on contextualised embeddings of the head and tail entity for predicting relationships. On the other hand, prompt-based strategies can be used to obtain embeddings that represent the relationship itself. The latter strategy is arguably more intuitive, but we found it to perform poorly in practice. Rather than suggesting that such relation embeddings are not useful, however, we found that they capture information that is highly complementary to what is captured by contextualised entity embeddings. Indeed, we considered a hybrid strategy, which substantially outperforms either of the two individual strategies, allowing us to improve the state of the art in each of the five considered benchmarks. Remarkably, we found that this remains true if we use entity embeddings from an off-the-shelf entity encoder. Finally, as a secondary contribution, we also found that coreference chains, which were used for training entity encoders in a self-supervised way by \citet{mtumbuka2023encore}, can be successfully leveraged for self-supervised training of relation encoders.

\section*{Limitations}
Our analysis in this paper was limited to the English language, and we only considered the setting of fully supervised sentence-level relation extraction. Since our focus was on learning representations (i.e.\ relation embeddings), it is not straightforward to transfer our findings to the zero-shot relation extraction setting (as we do not attempt to model the relation labels). 
We did not consider the use of LLMs for relation extraction in this paper. On the one hand, this is due to the fact that applying LLMs to this setting is not straightforward, especially when the set of candidate relation labels is large. Progress in this area has indeed been slow, as we highlighted in the related work section. Moreover, while LLMs can be used to extract symbolic representations (e.g.\ knowledge graph triples), they are often less suitable for learning embeddings. Traditionally, relation embeddings have primarily been used as an intermediate representations, before relation labels were predicted, and from this perspective, we may wonder whether relation embeddings are still needed in the LLM era. However, beyond acting as an intermediate representation, embeddings have a number of important advantages. They can, in principle, capture much more subtle distinctions than is possible with pre-defined discrete relation labels. As such, they are more suitable for modelling relational similarity (i.e.\ analogy), for instance.

% Only for English
% Only for sentence-level relation extraction
% LLMs may make relation encoders obsolete

%\section*{Ethics Statement}

\paragraph{Acknowledgements} This research was supported by EPSRC grant EP/W003309/1 and undertaken using the supercomputing facilities at Cardiff University operated by Advanced Research Computing at Cardiff (ARCCA) on behalf of the Cardiff Supercomputing Facility and the HPC Wales and Supercomputing Wales (SCW) projects. We acknowledge the support of the latter, which is part-funded by the European Regional Development Fund (ERDF) via the Welsh Government.

\bibliography{anthology,custom}

\appendix

%\newpage
\section{Training Details}

%\paragraph{Pre-training Strategies}
%Figures \ref{figHTpM} and \ref{figHTM} respectively illustrate the [H,T]+Mask and the [H,T,Mask] pre-training strategies. As explained in the main paper, once the encoders have been pre-trained, both strategies use the same approach: they fine-tune a relation classifier which takes as input the representations of the head and tail entities (i.e.\ the embeddings of the [E1] and [E2] tokens) as well as the embedding of the [MASK] token. However, in the case of the [H,T]+Mask strategy, pre-training involves two InfoNCE losses: one which ensures that the [H,T] representations are informative and one which ensures that the [MASK] embedding is informative. In contrast, for the [H,T,Mask] strategy, pre-training uses a single InfoNCE loss, which directly uses the concatenation of the [H,T] and Mask representations.

%\paragraph{Training Details}
We pre-train the model for 25 epochs and select the checkpoint with the minimum validation loss.
% \todo{Give details for pre-training step}  
For the fine-tuning step, we similarly train the model for 25 epochs and select the best checkpoint based on the validation set. We use the AdamW optimizer \cite{loshchilov2018decoupled} with a learning rate of $5e-4$ and a weight decay $\lambda$ of $1e-5$.
The temperature $\tau$ in the contrastive loss was set to 0.05. 
% For the relation classifier, we used 256-dimensional embeddings for the hidden layer.
%\frank{We used the size of the embeddings from the last hidden layer for the corresponding pre-trained relation encoder as the size for the hidden layer in the relation classifier.}

\section{Analysis}
In this section we provide some further analysis of the different relation embeddings. In particular, we make the following observations:
\begin{itemize}
\item By analysing model outputs, we find that the ``no relation'' label (from TACRED) is the most difficult label for all variants, where models are particularly prone to predicting some relationship even if none is expressed in the sentence. The [H,T]+Mask variant suffers considerably less from this problem, which helps to explain its outperformance over [H,T].
\item In a qualitative analysis of the model outputs, we provide further evidence that [H,T] focuses too much on the entity types. In particular, the model frequently predicts a relationship that holds for entities of the same type but is not expressed in the sentence.
\item We show that the predictions of the masked language model for the [MASK] token are largely meaningless, even though the contextualised representation of this token is useful for relation classification. This is expected, given that the model is not trained to produce meaningful verbalisations of the relationship.
\end{itemize}

% \frank{
% Notes:
% \begin{itemize}
%     \item When [H, T] is given entity types of person and country, it's likely to predict either the relation is residency or origin without taking full context into account. This is seen in the first, fourth, and fifth examples in Table~\ref{tabComparisonModelOutputs}.
%     \item When [H, T] is given entity types of person and organization, it's likely to predict either the relation is organisational affiliation  without taking full context into account. This is seen in the second example.
%     \item Deliberately included the last example in the table to show where [H, T] + Mask missed.
%     \item In the third example, [H, T]+ Mask is able to pick up a longer context between the entities involved and is able to predict the relations correctly.
% \end{itemize}
% }

% \paragraph{Comparison of the Confusion Matrices}
% % In Figures~\ref{fig:HTConfusionMatrix} and \ref{fig:HTMaskConfusionMatrix}, we show the confusion matrix for 5 randomly sampled relations and the ``no relation'' label, for the TACREV test set. For this analysis, we focus on the [H,T] and [H,T]+Mask models. We group any predictions of labels outside the sampled six classes as the ``other'' class. 
% % We can see that the ``no relation'' label is the most challenging label for both models: most errors arise when the models predict a relationship when there is none, or vice versa. We can also see, however, that the [H,T]+Mask consistently improves on [H,T] when it comes to modelling the ``no relation'' label. This supports the conclusions from our qualitative analysis.

\begin{figure*}
   \centering
   \includegraphics[width=0.7\textwidth]{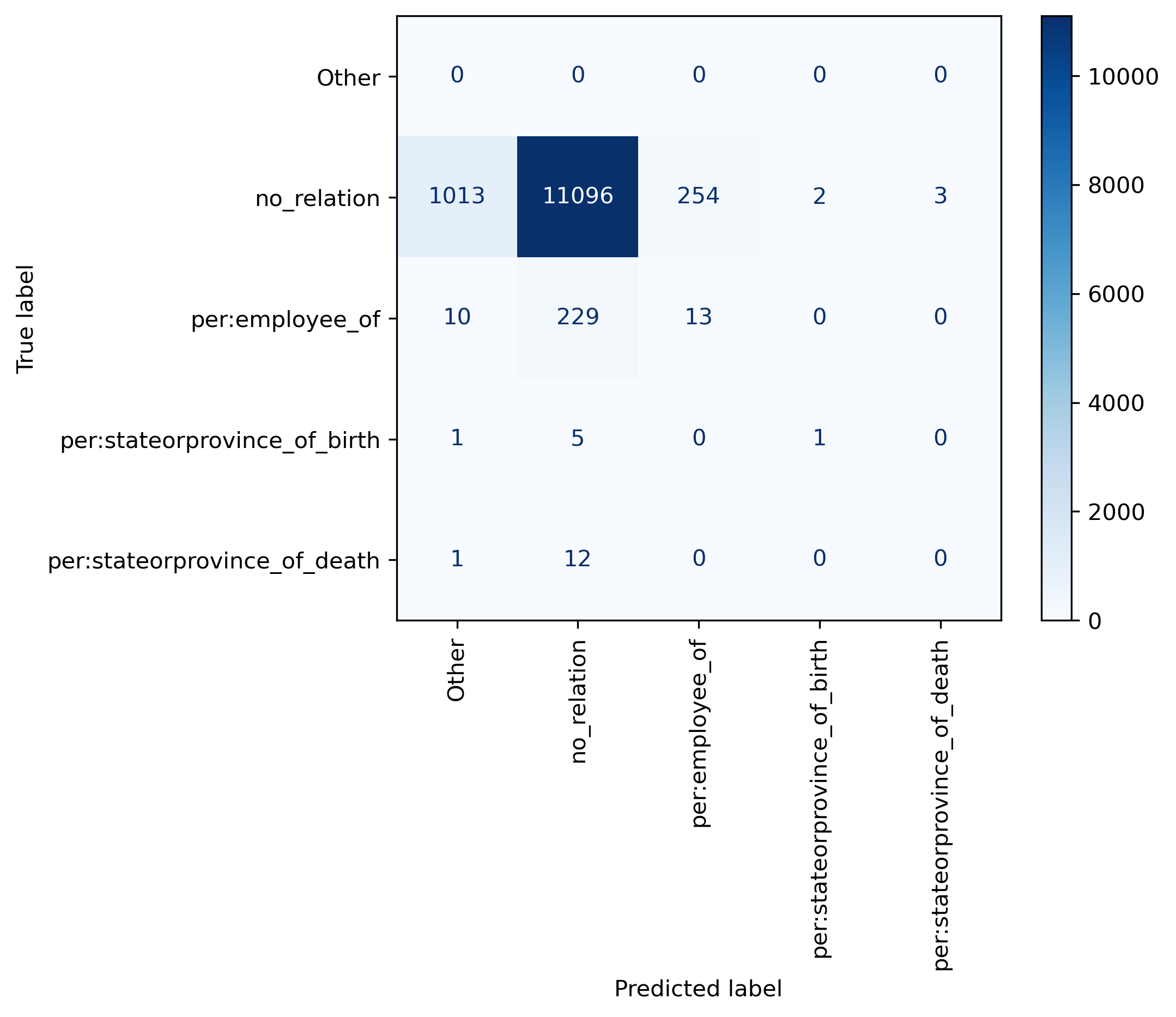}
   \caption{The confusion matrix for the Mask model on the TACREV test set.}
   \label{fig:MaskConfusionMatrix}
 \end{figure*}

\begin{figure*}
   \centering
   \includegraphics[width=0.7\textwidth]{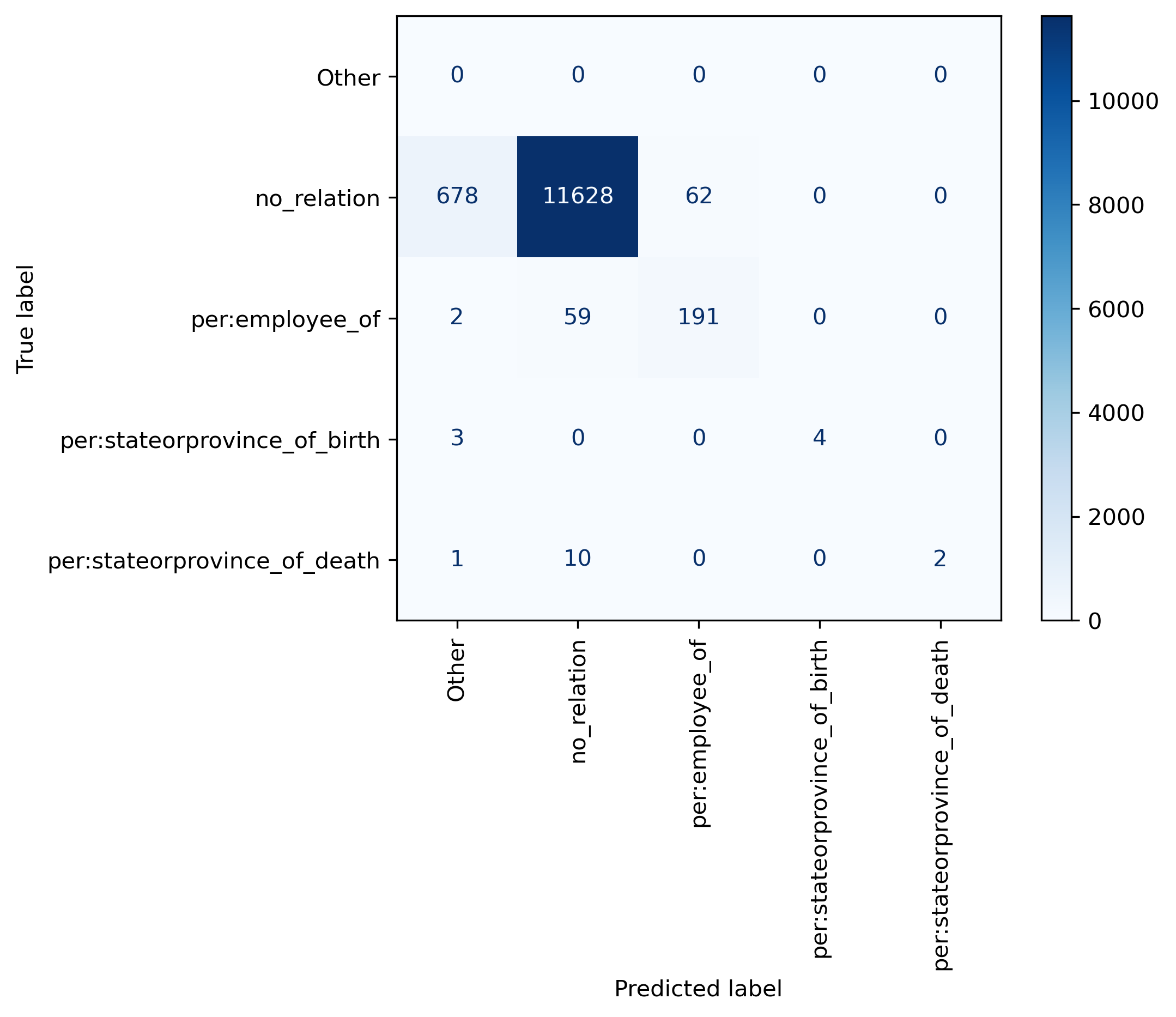}
   \caption{The confusion matrix for the [H,T] model on the TACREV test set.}
   \label{fig:HTConfusionMatrix}
 \end{figure*}

\begin{figure*}
   \centering
   \includegraphics[width=0.7\textwidth]{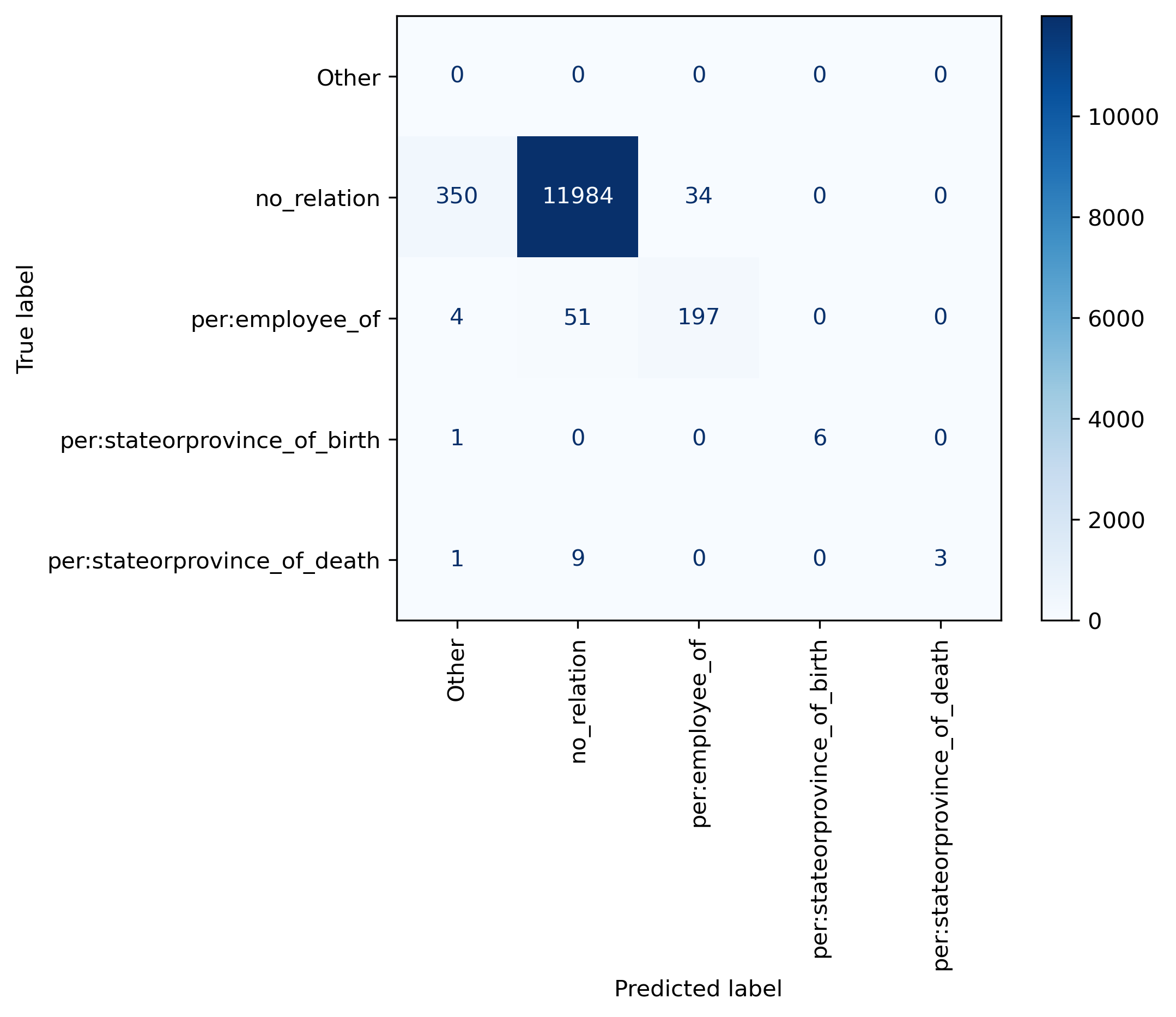}
   \caption{The confusion matrix for the [H,T]+ Mask model on the TACREV test set.}
   \label{fig:HTMaskConfusionMatrix}
 \end{figure*}

\paragraph{Confusion Matrices}
In Figures~\ref{fig:MaskConfusionMatrix},~\ref{fig:HTConfusionMatrix} and \ref{fig:HTMaskConfusionMatrix}, we show the confusion matrix for 5 randomly sampled relations and the ``no relation'' label, for the TACREV test set. For this analysis, we focus on the Mask, [H,T] and [H,T]+Mask models. We refer to any label predictions that fall outside of the six sampled classes as the ``other" class. We can clearly observe that the ``no relation" label is the most difficult label for all three models: most errors occur when the models predict a relationship while none exists, or vice versa. We can also see that the [H,T]+Mask model regularly outperforms [H,T], which in turn outperforms the Mask model when it comes to modelling the ``no relation" label. 

\begin{figure*}[t]
  \centering
  \includegraphics[width=0.9\textwidth]{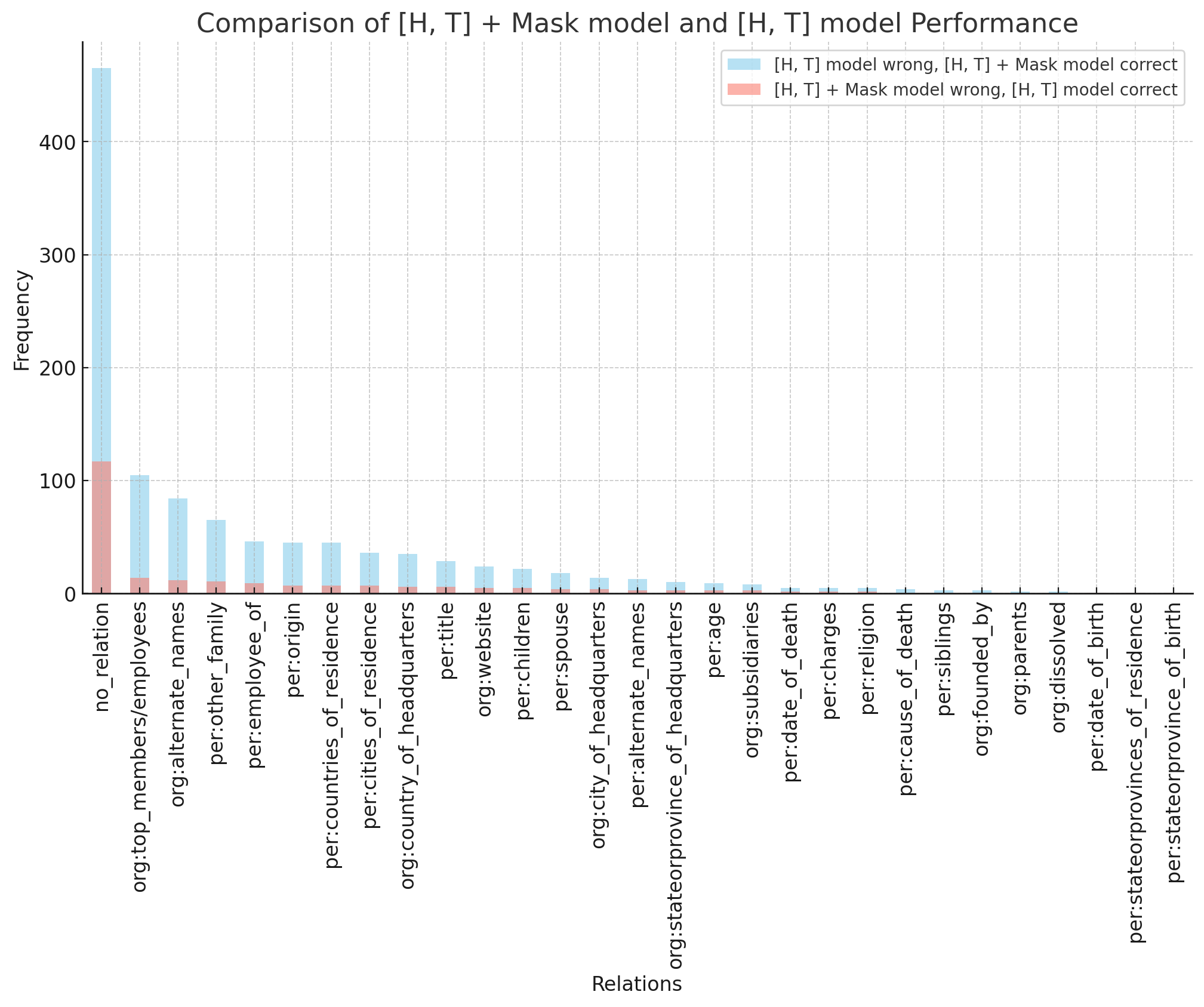}
  \caption{The comparison of the errors made by the [H,T] and [H,T]+Mask models on the TACREV test set.}
  \label{fig:predictionsComparison}
\end{figure*}

\paragraph{Error Comparison}
In Figure~\ref{fig:predictionsComparison}, we present the analysis of model performance that highlights instances where one model outperforms the other. Specifically, we focus on the [H,T] and [H,T]+mask models, showing how often one model makes a mistake while the other gets it right, for the TACREV test set. Compatible with our findings from the confusion matrices, the no ``relation'' label proves challenging for both methods. There are several instances where [H,T] makes a mistake while [H,T]+Mask does not, and vice versa, although [H,T]+Mask overall performs better. When we look at the other relation labels, however, we can see a clear pattern, as there are only very few instances where [H,T]+Mask makes a mistake without [H,T] also making the same mistake. This shows that the improvement of [H,T]+Mask over [H,T] is highly consistent.

\paragraph{Qualitative Analysis of Model Outputs} 
Overall, we find that around  70\% of the errors of the [H,T] model are ``no relation'' misclassifications, where most of the remaining mistakes arise because the model confuses relations between entities of the same type, including:
\begin{itemize}
\item family relationships such as ``per:siblings'' and ``per:children'';
\item relationships linking people to geographic regions, such as ``per:state or province of death'' and ``per:state or province of residence'';
\item relationships linking people to organisations, such as ``org:founded by'' and ``org:top members/employees''.
\end{itemize}
We provide several examples of both types of errors. In particular, Table \ref{tabComparisonModelOutputsAppendixNoRel} shows cases where [H,T] mistakenly predicts the ``no relation'' label, while Table \ref{tabComparisonModelOutputsAppendixConfused} focuses on examples where the [H,T] model confuses the target relation with a relation between entities of the same type. 

The Mask model performs worse overall, and there is less of a pattern in the types of errors it makes. Similar to the other models, it is also prone to incorrectly predicting ``no relation''. Several examples of this can be seen in Tables \ref{tabComparisonModelOutputsAppendixNoRel} and \ref{tabComparisonModelOutputsAppendixConfused}. One particular weakness of Mask is that it sometimes fails to correctly predict the direction of a relationship, confusing the target relation with its inverse. This can be seen, for instance, in the last example of Table \ref{tabComparisonModelOutputsAppendixConfused}, where Mask confuses ``per:children'' with ``per:parents''. Another example can be found in Table \ref{tabComparisonModelOutputs} (fourth instance), where the Mask model confused ``org:top members/employees'' with ``per:employee of''. As we highlighted in our qualitative analysis in the main paper, Mask also makes mistakes because it fails to take into account the semantic types of the entities.

\paragraph{Encoder Predictions for the [MASK] token}
 In Table~\ref{tabMaskPredictions}, we present examples of the tokens which are predicted by the language model for the [MASK] position of the appended relation prompts. Specifically, we use our pre-trained \textit{[H, T] + Mask} encoder that was initialised using RoBERTa-large. We consider the top five tokens directly predicted for the [MASK] positions by the encoder. We compare these predictions with the predictions from our full \textit{[H, T] + Mask} model, which comes with a classifier on top of the encoder. As can be seen, the token predictions do not adequately capture the relationships that are expressed in the given sentences. This is to be expected, since the InfoNCE loss which is used during pre-training encourages the embeddings of sentences that express the same relationship to be similar, but these vectors are no longer aligned with the tokens from the encoder's vocabulary. %Additionally, the gold relation labels in the benchmark datasets are distinct from what is expressed in the \acp{LLM}' vocabularies. 
 This illustrates the need for a classifier that maps the embeddings of the [MASK] token in the relation prompt to dataset-specific relation labels. %This is why the full \textit{[H, T] + Mask} model is able to predict the relation labels that are similar to the gold labels in the datasets.

 \begin{table*}[t]
    \centering
    \footnotesize
    \setlength\tabcolsep{4pt}
    \begin{tabular}{p{265pt}p{170pt}}
        \toprule
         \textbf{Sentence} 
         & \textbf{Label}\\ 
         \midrule
         ``The current attempt to restore the commission was masterminded by a suspected \textbf{mobster}, \textbf{Matteo Messina Denaro}, who is among a handful of people vying to replace Provenzano," police said.
         & Gold: per:title \newline Mask: no relation \newline [H,T]: no relation \newline [H,T]+Mask: per:title\\

         \midrule
         She ended up leaving Iraq under the threat of losing \textbf{her} job and returning home to \textbf{Texas} to seek medical and psychiatric treatment for post traumatic stress syndrome.
         & Gold: per:state or provinces of residence \newline Mask: no relation \newline [H,T]: no relation \newline [H,T]+Mask: per:state or provinces of residence\\
\midrule
 Wen was tried with \textbf{his} wife, \textbf{Zhou Xiaoya}, and three former Chongqing police associates all of whom received jail sentences of up to 20 years.
         & Gold: per:spouse \newline Mask: no relation \newline [H,T]: no relation \newline [H,T]+Mask: per:spouse\\ 

    \midrule
         Knox's father, Curt Knox, said \textbf{his} daughter looked ``confident in what \textbf{she} wants to say."
         & Gold: per:children \newline Mask: no relation \newline [H,T]: no relation \newline [H,T]+Mask: per:children\\       
          \midrule          
         \textbf{Her} mother, 60-year-old \textbf{Claudie Mamane}, tried to jump from the van while it was still moving and injured her arm.
         & Gold: per:parents \newline Mask: no relation \newline 
         [H,T]: no relation 
         \newline [H,T]+Mask: per:parents\\
         \midrule  
         \textbf{He} is also survived by his parents and a sister, Karen Lange, of Washington, and a brother, \textbf{Adam Lange}, of St. Louis.
         & Gold: per:siblings  \newline Mask: per:parents \newline 
         [H,T]: no relation 
         \newline [H,T]+Mask: per:siblings\\
         \midrule         
         After more than 20 years of wearing the same long hairstyle, \textbf{silver-tressed Ponce Kiah Marchelle Heloise Cruse Evans}, known simply as \textbf{Heloise}, has a new 'do for the New Year.
         & Gold: per:alternate names  \newline Mask: no relation \newline 
         [H,T]: no relation 
         \newline [H,T]+Mask: per:alternate names\\
    \bottomrule
    \end{tabular}
    \caption{Comparison of the TACREV test set predictions from the Mask, [H, T] and [H, T] + Mask models that were initialised using RoBERTa-large, focusing on examples where [H,T] incorrectly predicts ``no relation''.}
    \label{tabComparisonModelOutputsAppendixNoRel}
\end{table*}

 \begin{table*}[t]
    \centering
    \footnotesize
    \setlength\tabcolsep{4pt}
    \begin{tabular}{p{285pt}p{150pt}}
        \toprule
         \textbf{Sentence} 
         & \textbf{Label}\\ 
            \midrule
         \textbf{A man} who shot and killed three women in a \textbf{Pennsylvania} health club, then himself, apparently blogged his rage-filled preparations --with the final chilling entry announcing the ``big day" of the massacre.
         & Gold: per:state or province of death \newline Mask: no relation \newline [H,T]: per:state or provinces of residence \newline [H,T]+Mask: per:state or province of death\\

\midrule
  \textbf{Her} brother-in-law, \textbf{Wen}, served as a top Chongqing police official for 16 years before taking over the city's judiciary.
         & Gold: per:other family \newline Mask: per:siblings \newline [H,T]: per:siblings \newline [H,T]+Mask: per:other family\\ 

         \midrule
 %        \textbf{Gross}, a 60-year-old native of Potomac, \textbf{Maryland}, was working in Cuba for a firm contracted by USAID when he was arrested as a suspected spy in Havana on Dec 3.
  %       & Gold: per:state or province of birth \newline Mask: no relation \newline [H,T]: per:state or province of birth \newline [H,T]+Mask: per:state or provinces of residence\\
   %      \midrule
         Robert Holden, deputy director at \textbf{the National Congress of American Indians}, said ``\textbf{the Washington DC-based group} is hopeful that the use of secured cards could be expanded to allow tribal members to travel abroad."
         & Gold: org:state or province of headquarters \newline Mask: no relation \newline [H,T]: per:state or provinces of residence\newline [H,T]+Mask: per:state or provinces of residence\\
         \midrule
         \textbf{Countrywide Financial Corp} co-founder \textbf{Angelo Mozilo} retired amidst scandal and investigation.
         & Gold: org:founded by \newline Mask: no relation \newline 
         [H,T]: org:top members/employees 
         \newline [H,T]+Mask: org:founded by\\
         \midrule          
         \textbf{Lt. Assaf Ramon}, the son of Israel's first astronaut, \textbf{Col. Ilan Ramon}, who died in the space shuttle Columbia disaster in 2003, was killed Sunday when an F16-A plane he was piloting crashed in the hills south of Hebron in the West Bank.
         & Gold: per:children  \newline Mask: per:parents \newline 
         [H,T]: per:siblings 
         \newline [H,T]+Mask: per:children\\        
    \bottomrule
    \end{tabular}
    \caption{Comparison of the TACREV test set predictions from the Mask, [H, T] and [H, T] + Mask models that were initialised using RoBERTa-large, focusing on examples where [H,T] confuses the ground truth with a different relation between entities of the same type.}
    \label{tabComparisonModelOutputsAppendixConfused}
\end{table*}

\begin{table*}[t]
    \centering
    \footnotesize
    \setlength\tabcolsep{4pt}
    \begin{tabular}{p{235pt}p{200pt}}
        \toprule
         \textbf{Sentence} 
         & \textbf{Label}\\ 
         \midrule
         \textbf{Asia Bibi} was sentenced to hang in Pakistan's central province of \textbf{Punjab} earlier this month after being accused of insulting the Prophet Mohammed in 2009.
         & Gold: no relation
         \newline
         Token predictions: strained, complicated, tense, fraught, complex.
         \newline [H,T]+Mask: no relation
         \\

         \midrule
         \textbf{Benjamin Chertoff} is the Editor in Chief of Popular Mechanics magazine as well as the cousin of the Director of \textbf{Homeland Security}, Michael Chertoff.
         & Gold: no relation 
         \newline
         Token predictions: unclear, unknown, complicated, complex, clear.
         \newline [H,T]+Mask: no relation
         \\
         \midrule
         ``The current attempt to restore the commission was masterminded by a suspected \textbf{mobster}, \textbf{Matteo Messina Denaro}, who is among a handful of people vying to replace Provenzano," police said.
         & Gold: per:title 
         \newline
         Token predictions: unclear, unknown, known, murky, complex.
         \newline
         [H,T]+Mask: per:title
         \\

         \midrule
         \textbf{A man} who shot and killed three women in a \textbf{Pennsylvania} health club, then himself, apparently blogged his rage-filled preparations --with the final chilling entry announcing the ``big day" of the massacre.
         & Gold: per:state or province of death 
         \newline
         Token predictions: unclear, unknown, clear, murky, chilling.
          \newline 
          [H,T]+Mask: per:state or province of death
         \\

         \midrule
         \textbf{Her} brother-in-law, \textbf{Wen}, served as a top Chongqing police official for 16 years before taking over the city's judiciary.
         & Gold: per:other family 
         \newline
         Token predictions: unclear, complicated, unknown, strained, complex.
         \newline 
         [H,T]+Mask: per:other family
         \\

         \midrule
         Robert Holden, deputy director at \textbf{the National Congress of American Indians}, said ``\textbf{the Washington DC-based group} is hopeful that the use of secured cards could be expanded to allow tribal members to travel abroad."
         & Gold: org:state or province of headquarters 
         \newline
         Token predictions: unclear, confidential, unknown, complex, complicated.
         \newline 
         [H,T]+Mask: per:state or provinces of residence\\
                
    \bottomrule
    \end{tabular}
    \caption{Comparison of the tokens predicted for the [MASK] position when using the pre-trained [H, T] + Mask encoder that was initialised using RoBERTa-large. Examples were selected from the TACREV test set. The token predictions are arranged in decreasing order of confidence (score).}
    \label{tabMaskPredictions}
\end{table*}

%****************************************************************
\section{Ablations and Additional Experiments}

\paragraph{Perfomance of the [CLS] token} 
%\frank{In addition, we provide a baseline in which we just consider the embedding of the [CLS] token to predict the relation type. We consider just using this token because it encodes contextual information in a given sentence~\cite{devlin-etal-2019-bert}. In this baseline, we merely train the classifier and leave the encoder frozen.}
The intuition behind the Mask strategy is that the embedding of the [MASK] token captures the relational context. Another possible approach is to use the [CLS] token for this purpose. Table \ref{tab:cls_token_performance} analyses a number of variants that are based on this idea. Specifically, we consider the [CLS] token on its own, as well as variants of the [H,T,Mask] and [H,T]+Mask strategies where the role of the [MASK] token is replaced by the [CLS] token. These latter two strategies are referred to as [H,T,CLS] and [H,T]+CLS. For this analysis, we have used TACREV with standard pre-training. 
Overall, we can draw the same conclusions as for the variants with the [MASK] token, in particular when it comes to comparing [H,T,CLS] with [H,T]+CLS. However, the [CLS] variants yield results which are somewhat worse than the counterparts based on [MASK] (with the exception of the case where [CLS] is used on its own). This justifies the use of a prompt with [MASK].

\paragraph{Impact of Masking Approaches} 
\citet{peng-etal-2020-learning} highlighted the importance of masking entity spans during pre-training, with some probability, to prevent the model from relying too much on the entity names themselves. Indeed, models which rely too much on the entity names are prone to learning shortcuts which hamper generalisation, a problem which is sometimes referred to as entity bias \cite{wang-etal-2022-rely}. Inspired by these works, we further investigate the impact of different masking strategies during pre-training. 
Specifically, we look at the following four scenarios in Table \ref{tab:masking_strategies}. First, in the strategy labelled \textit{No masking}, we do not mask any tokens in the input corpus. Second, in the case of \textit{Mask entity spans}, for each entity, we mask the entire entity span with 15\% probability. For the variant labelled \textit{Mask entity span heads}, we merely mask out the syntactic heads of entities, for 15\% of the entity spans. We find the head word using the SpaCy dependency 
parser\footnote{\url{https://spacy.io/api/dependencyparser}}. This is motivated by the idea that the syntactic head is most likely to reveal the entity type. Masking this head was found to by beneficial when pre-training entity encoders for this reason \cite{mtumbuka2023encore}.
Finally, in the case of \textit{Random tokens}, we randomly mask 15\% of the tokens in the training corpus. These tokens include both tokens from entity spans and non-entity tokens. This is the strategy that we have used for the main experiments. We can see that masking random tokens gives us the best results.
%Table~\ref{tab:masking_strategies} presents the results on the performance of different masking approaches. We can see that the best performance comes from the setting where we randomly mask out 15\% of the tokens.

\paragraph{Comparison of Language Models}
%\paragraph{Comparison of Different Language Models}

In Table \ref{tabComparisonLMsHTMaskvariants}, we compare the performance of \texttt{roberta-large}, which we have been using for our main experiments, with \texttt{bert-base-uncased}\footnote{\url{https://huggingface.co/docs/transformers/model_doc/bert}},  
\texttt{bert-large-uncased} and
 \texttt{albert-xxlarge-v1}\footnote{\url{https://huggingface.co/docs/transformers/model_doc/albert}}. 
 % \todo{Discuss ...}
 Unsurprisingly, we can see that \texttt{bert-large-uncased} outperforms \texttt{bert-base-uncased}. Furthermore, across all datasets, \texttt{roberta-large} outperforms \texttt{bert-large-uncased}, but the best results are obtained by \texttt{albert-xxlarge-v1}. This is consistent with the findings from \citet{zhong-chen-2021-frustratingly}. The main advantage of \texttt{albert-xxlarge-v1} is that it uses 4096-dimensional embeddings, compared to 1024-dimensional embeddings for \texttt{bert-large-uncased} and \texttt{roberta-large}, despite being smaller than the latter two models (due to parameter sharing across layers).  
 %\frank{ In Appendix~\ref{app:diff_backbones}, we further deepen our analysis by investigating the comparative analysis of different relation embedding strategies on a range of backbones. The other relation embedding strategies continue to perform worse when compared to the [H, T] + Mask strategy, whose results are presented in Table~\ref{tabComparisonLMs}, but overall, the results show a consistent pattern. \texttt{Roberta-large} comes in second, \texttt{bert-large-uncased} is third, and \texttt{bert-base-uncased} is last in terms of performance. \texttt{Albert-xxlarge-v1} always performs the best.
% }
As can be seen, for all language models, the main conclusions remain consistent with those reported in the main paper. For instance, we consistently find that the Mask strategy in isolation yields the lowest performance, that [H,T,Mask] improves on [H,T] and that [H,T]+Mask achieves the best results. This further supports our main findings about the complementarity of the [H,T] and Mask representations. 

 \paragraph{Few-shot Relation Classification}
To complement our main results, we have carried out an evaluation in the few-shot setting, where only a few training examples per relation are available.
We employ three commonly-used few-shot learning settings on the family of TACRED datasets: 4 training examples per relation (representing approximately 1\% of the full TACRED training set), 16 training examples per relation (approximately 5\%), and 32 examples per relation (approximately 10\%)~\cite{sainz-etal-2021-label,lu-etal-2022-summarization}. We compare our models with the methods from \citet{sainz-etal-2021-label} and \citet{lu-etal-2022-summarization}, which were specifically designed for the few-shot setting. In particular, \citet{sainz-etal-2021-label} relied on pre-trained NLI models to solve relation extraction in the few-shot setting. Note that their approach relies on manually constructed verbalisations of the relations, which essentially provides an additional supervision signal. They reported results for NLI models based on RoBERTa-large (shown as $\text{NLI}_{\text{RoBERTa}}$) on DeBERTa-v2xxlarge (shown as $\text{NLI}_{\text{DeBERTa}}$). \citet{lu-etal-2022-summarization} treat few-shot relation extraction as a summarisation task, relying on a pretrained PEGASUS-large abstractive summarisation model (shown as $\text{SURE}_{\text{PEGASUS}}$). They also rely on manually constructed verbalisations.

As can be seen in Table~\ref{tabFewShot}, the [H,T]+Mask model consistently outperforms $\text{SURE}_{\text{PEGASUS}}$. Furthermore, [H,T]+Mask outperforms $\text{NLI}_{\text{RoBERTa}}$ in all cases apart from the 1\% setting, and $\text{NLI}_{\text{DeBERTa}}$ for the 10\% and 100\% configurations. This is remarkable, given that these baselines were specifically designed for the few-shot setting, rely on extensive pre-training, and in the case of $\text{NLI}_{\text{DeBERTa}}$ and $\text{SURE}_{\text{PEGASUS}}$ rely on much larger LMs. When it comes to the relative performance of the different variants that are considered in this paper, our overall findings are similar as for the main experiments. For instance, [H,T]+Mask and Encore+Mask achieve the best results, the performance of EnCore+Mask is again similar to that of [H,T]+Mask, and pre-training on Gigaword consistently improves the results. The performance of Mask and [H,T] is particularly poor in the few-shot setting, and combining these two types of representations has a very big impact here. For instance, in the setting with 1\% of the training data with standard pre-training, the performance increases from 19.7\% for [H,T] to 48.4\% for [H,T,Mask] and 53.0\% for [H,T]+Mask.

\begin{table}[t]
 	\begin{center}
 		\footnotesize
 			\centering     
     	    \begin{tabular}{lc}
     		\toprule    
            &\textbf{TACREV}\\
       \midrule     
        Mask & 22.9 \\
        \text{[H, T, Mask]}  & 73.8\\
        \text{[H, T] + Mask} & 84.6\\
       \midrule     
        CLS & 25.4 \\
        \text{[H, T, CLS]}  & 71.6\\
        \text{[H, T] + CLS} &  79.2\\
        \bottomrule                                
         		\end{tabular}
     	\caption{Evaluation of strategies using the [CLS] token on TACREV. For this analysis, we have used RoBERTa-large with the standard pre-training strategy.}\label{tab:cls_token_performance}%
     	\end{center}
     \end{table}

\begin{table}
 	\begin{center}
 		\footnotesize
 			\centering             
     	    \begin{tabular}{lc}
     		\toprule
            & \textbf{TACREV} \\
       \midrule       
        No masking & 58.7 \\
        Mask entity spans &  73.8\\
        Mask entity span heads  & 79.4\\
        Mask random tokens &  84.6\\        
        \bottomrule                                
         		\end{tabular}
     	\caption{Evaluation of different masking strategies. For this analysis, we have used the \text{[H, T] + Mask} approach with RoBERTa-large and standard pre-training.}
      \label{tab:masking_strategies}%
     	\end{center}
     \end{table}

\begin{table}[t]
\centering
\footnotesize
\begin{tabular}{llccccc}
\toprule
& & & \rotatebox{90}{\textbf{BERT-base}} & \rotatebox{90}{\textbf{BERT-large}} & \rotatebox{90}{\textbf{RoBERTa-large}} & \rotatebox{90}{\textbf{ALBERT-xxl}} \\
\midrule
\parbox[t]{2mm}{\multirow{6}{*}{\rotatebox[origin=c]{90}{\textbf{Mask}}}} & TACRED & F1 & 20.9 & 22.2 & 23.3 & \textbf{25.1} \\
& TACREV & F1 & 20.6 & 21.7 & 22.9 & \textbf{24.8} \\
& Re-TACRED & F1 & 21.3 & 22.1 & 23.2 & \textbf{25.4} \\
& Wiki-WD & F1 & 44.9 & 46.2 & 48.9 & \textbf{51.3} \\
& NYT-FB & P@10 & 48.2 & 49.5 & 52.3 & \textbf{53.7} \\
& NYT-FB & P@30 & 47.6 & 48.2 & 49.7 & \textbf{51.2} \\
\midrule 
\parbox[t]{2mm}{\multirow{6}{*}{\rotatebox[origin=c]{90}{\textbf{[H,T]}}}}& TACRED & F1 & 58.6 & 59.3 & 61.4 & \textbf{62.9} \\
& TACREV & F1 & 59.1 & 60.7 & 63.6 & \textbf{64.3} \\
& Re-TACRED & F1 & 67.4 & 69.1 & 72.3 & \textbf{73.5} \\
& Wiki-WD & F1 & 72.4 & 73.1 & 75.2 & \textbf{76.9} \\
& NYT-FB & P@10 & 75.2 & 76.7 & 79.6 & \textbf{80.7} \\
& NYT-FB & P@30 & 73.9 & 75.2 & 78.5 & \textbf{79.3} \\
\midrule
\parbox[t]{2mm}{\multirow{6}{*}{\rotatebox[origin=c]{90}{\textbf{[H,T,Mask]}}}} & TACRED & F1 & 70.1 & 71.6 & 73.0 & \textbf{74.7} \\
& TACREV & F1 & 70.8 & 71.1 & 73.8 & \textbf{75.2} \\
& Re-TACRED & F1 & 76.3 & 78.4 & 81.9 & \textbf{84.7} \\
& Wiki-WD & F1 & 76.2 & 78.7 & 81.7 & \textbf{82.4} \\
& NYT-FB & P@10 & 80.5 & 84.3 & 88.4 & \textbf{90.2} \\
& NYT-FB & P@30 & 79.9 & 82.7 & 86.1 & \textbf{89.4} \\
\midrule
\parbox[t]{2mm}{\multirow{6}{*}{\rotatebox[origin=c]{90}{\textbf{[H,T] + Mask}}}} & TACRED & F1 & 74.8 & 75.1 & 78.5 & \textbf{78.9} \\
& TACREV & F1 & 75.6 & 77.3 & 84.6 & \textbf{83.9} \\
& Re-TACRED & F1 & 86.4 & 87.9 & 91.9 & \textbf{92.3} \\
& Wiki-WD & F1 & 79.7 & 84.9 & 89.6 & \textbf{89.8} \\
& NYT-FB & P@10 & 93.9 & 94.1 & 94.7 & \textbf{94.9} \\
& NYT-FB & P@30 & 93.4 & 92.5 & 93.1 & \textbf{93.8} \\
\bottomrule
\end{tabular}
\caption{Comparison of different language models, using different relation embedding strategies under standard pre-training.}
\label{tabComparisonLMsHTMaskvariants}
\end{table}

 \begin{table}[!t]
\centering
\footnotesize
\begin{tabular}{lcccc}
\toprule
& \multicolumn{4}{c}{\textbf{F1}}\\
\cmidrule(lr){2-5}
& \textbf{1\%} & \textbf{5\%} & \textbf{10\%} & \textbf{100\%} \\

\midrule
\multicolumn{5}{c}{\textit{Baselines}}\\
\midrule
 SpanBERT$^{\dagger}$ & 0.0 & 28.8 & 1.6 & 70.8 \\
 RoBERTa$^{\dagger}$  & 7.7 & 41.8 & 55.1 & 71.3 \\
 LUKE$^{\dagger}$ & 17.0 & 51.6 & 60.6 & 72.0 \\
 $\text{NLI}_{\text{RoBERTa}}^{\dagger}$ & 56.1 & 64.1 & 67.8 & 71.0 \\
 $\text{NLI}_{\text{DeBERTa}}^{\dagger}$ &  \textbf{63.7} & \textbf{69.0}  & 67.9 & 73.9 \\
 $\text{SURE}_{\text{PEGASUS}}^*$  &  52.0 & 64.9 & 70.7 & 75.1 \\
\midrule
\multicolumn{5}{c}{\textit{Standard pre-training}}\\
\midrule
 Mask & 8.3 & 11.6 & 15.0 & 23.3 \\
 \text{[H, T]} & 19.7 & 27.5 & 33.9 & 61.4\\
 \text{[H, T, Mask]} & 48.4 & 54.6 & 61.3 & 73.0 \\
 \text{[H, T] + Mask} &  53.0  & 65.9  &  \textbf{71.7} &  78.5 \\
 \text{EnCore + Mask} &  52.9 & 65.8  & 71.6 &  \textbf{78.9} \\
 \midrule
\multicolumn{5}{c}{\textit{Gigaword pre-training}}\\
\midrule
 Mask & 9.9 & 13.1 & 18.3 & 24.6 \\
 \text{[H, T]} & 21.1 & 28.7 & 34.3 & 63.2 \\
 \text{[H, T, Mask]} & 49.2 & 55.1 & 62.1 & 74.4\\
 \text{[H, T] + Mask} &  53.5  & 66.1  &  71.9 &  78.5 \\
 \text{EnCore + Mask} &  53.4 & 66.2  & 72.1 &  79.1 \\
\bottomrule
\end{tabular}
\caption{Few-shot scenario results on TACRED with 1\%, 5\%, 10\% and 100\% of training data. [H, T] + Mask was initialised using RoBERTa-large and standard pre-training on the reduced training sets are used for these experiments. The results for models marked with ${\dagger}$ were taken from \citet{sainz-etal-2021-label}, whereas those marked with $*$ were taken from \citet{lu-etal-2022-summarization}.}
\label{tabFewShot}
\end{table}

% Acronyms
\begin{acronym}
    \acro{BERT}{bidirectional encoder representations from transformers}
    \acro{KG}{knowledge graph}
    \acro{LLM}{large language lodel}
    \acro{LM}{language model}
    \acro{MLM}{masked language modelling}
    \acro{PLM}{pre-trained language model}
    \acro{RE}{relarion extraction}
\end{acronym}

\end{document}